\documentclass[11pt,a4paper]{article}
\usepackage[hyperref]{eacl2021paper}
\usepackage{times}
\usepackage{latexsym}

\usepackage{microtype}
\usepackage{graphicx}
\aclfinalcopy 

\usepackage{titlesec}

\titleformat{\paragraph}
{\normalfont\normalsize\bfseries}{\theparagraph}{1em}{}
\titlespacing*{\paragraph}
{0pt}{3.25ex plus 1ex minus .2ex}{1.5ex plus .2ex}

\title{Orthogonal Attention: A Cloze-Style Approach to Negation Scope Resolution}

\author{Aditya Khandelwal \\
  College of Engineering Pune\\ Pune, India \\
  \texttt{khandelwalar16.comp@coep.ac.in} \\\And
  Vahida Attar \\
  College of Engineering Pune\\ Pune, India \\
  \texttt{vahida.comp@coep.ac.in} \\}

\date{}

\begin{document}
\maketitle
\begin{abstract}

Negation Scope Resolution is an extensively researched problem, which is used to locate the words affected by a negation cue in a sentence. Recent works have shown that simply finetuning transformer-based architectures yield state-of-the-art results on this task. In this work, we look at Negation Scope Resolution as a Cloze-Style task, with the sentence as the Context and the cue words as the Query. We also introduce a novel Cloze-Style Attention mechanism called Orthogonal Attention, which is inspired by Self Attention. First, we propose a framework for developing Orthogonal Attention variants, and then propose 4 Orthogonal Attention variants: OA-C, OA-CA, OA-EM, and OA-EMB. Using these Orthogonal Attention layers on top of an XLNet backbone, we outperform the finetuned XLNet state-of-the-art for Negation Scope Resolution, achieving the best results to date on all 4 datasets we experiment with: BioScope Abstracts, BioScope Full Papers, SFU Review Corpus and the *sem 2012 Dataset (Sherlock). 
\end{abstract}

\section{Introduction}

Negation Scope Resolution involves finding the words in a sentence whose meaning was affected by the use of a negation cue (a word that expresses negation). Consider the following examples:
 \begin{enumerate}
     \item This place is[n’t] \emph{familiar.}
     \item I do [not] \emph{know the answer.}
     \item I am [neither] \emph{a saint} [nor] \emph{a sinner.} 
 \end{enumerate}
The words enclosed in square brackets are the negation cues, and the words in italics are their corresponding scopes. As we can see, negation cues can be of multiple types: an affix (1), a single word cue (2) and a multi-word cue (3). A sentence can also have multiple cue words, each of which can have different scopes. Hence, the input to any system performing Negation Scope Resolution is the sentence-negation cue pair. 

Approaches to solve the task of negation scope resolution have varied significantly over the years, ranging from simple rule based systems to BiLSTM and CRF classifiers. To represent the cue word input, traditionally, these methods utilized either cue dependent hand-crafted features (for CRF classifiers), or an additional binary input vector representing the cue words in the sentence. More recently though, \citet{khandelwal-sawant:2020:LREC} and \citet{britto2020resolving} used transformer-based architectures to address the task, and represented the cue words in the sentence via a preprocessing strategy: by augmenting the input sentence with a special token which is added before each cue word to tell the system that the following word is a cue word.

In this paper, we propose a novel approach to solving the problem of Negation Scope resolution: by viewing it as a Cloze-Style task, where the sentence is used as the Context input and the cue words are used as the Query input. We also develop a novel Cloze-Style Attention mechanism called Orthogonal Attention, which uses the key-query-value structure used in Self-Attention \cite{Vaswani2017AttentionIA}.

Cloze-Style Question Answering (and Machine Reading Comprehension) are 2 classes of problems that involve using 2 distinct inputs to produce the output desired. In most cases, the input is a query-context pair $\langle Q,C\rangle$ from which an answer $\langle A\rangle$ is generated. This is akin to posing a Question (Query) over a paragraph containing information (Context). The format of the answer can vary significantly, from pointing to a part of the context that contains the answer, as in SQuAD\cite{squad}, to filling in the blanks of the Query using relevant information from the Context. Thus, these tasks require modelling the interaction between the query and the context, to produce the corresponding answer. Such tasks became popular after the release of the CNN and Daily News Datasets in \citet{DBLP:journals/corr/HermannKGEKSB15}.

Given the success of Attention Mechanisms for traditional Natural Language Processing tasks (like translation), researchers have also developed Attention mechanisms to address Cloze-Style tasks (by using them to model the interaction between the Query and Context). A few examples of such Attention mechanisms include: Attention Sum Readers \cite{DBLP:journals/corr/KadlecSBK16}, Attention over Attention \cite{DBLP:journals/corr/CuiCWWLH16}, Gated-Attention  \cite{DBLP:journals/corr/DhingraLCS16}, and Dynamic Coattention Networks \cite{DBLP:journals/corr/XiongZS16}. Most attention mechanisms used a similar approach: compute a score for each query word-context word pair, and use that score to perform some form of a weighted summation of certain vectors to generate context-aware representations of the query and query-aware representations of the context. We review a few such attention mechanisms in Section 2.

The novel Cloze-Style Attention mechanism we propose is called Orthogonal Attention. We propose a framework to develop Orthogonal Attention variants, and propose 4 such variants (OA-C OA-CA, OA-EM, OA-EMB). The exact specifications are covered in Section 3, but at it's core, it uses the key-query-value structure and multiheaded structure found in Self Attention \cite{Vaswani2017AttentionIA}.

These layers are then appended to the current state-of-the-art architecture for Negation Scope Resolution (XLNet-base), and we observe that adding Orthogonal Attention layers to the architectures yield improvements over the current state-of-the-art architectures.

\section{Literature Review}

\subsection{Attention for Cloze-Style Tasks}
\citet{DBLP:journals/corr/KadlecSBK16} proposed Attention Sum Reader, wherein they performed a dot product between the question embedding and each context word embedding to get probability scores per word of the context. 


\citet{DBLP:journals/corr/DhingraLCS16} proposed a Gated Attention Reader Network to compute a query-aware representation of the context words. They used a multi-hop architecture with k hops (layers), where each layer incrementally infused the context embeddings with information from the query embeddings. They used a “Gated-Attention Module” for each word of the context, which performs attention over the query words using the context word followed by an element-wise multiplication between the summarized query and the context word. Each layer computes a different query representation using a separate BiGRU, and the output of the Gated-Attention Modules is also passed through a BiGRU before being fed to the next query.

\citet{DBLP:journals/corr/SeoKFH16} proposed Bidirectional Attention Flow (BiDAF), wherein they first computed a similarity matrix between the context words and the query words, which was then used to compute context-aware query representations and query-aware context representations. The context-to-query attention was modelled as an attention over the context representation using query tokens, where the attention weights came from the similarity matrix. The query-to-context attention produces a weighted sum of the query words, using the attention weights as the maximum across the columns of the similarity matrix. Finally, all the generated matrices and the original context embeddings were concatenated to generate the query-aware representation of each word.


\citet{wang-etal-2017-gated} introduced a Gated Self-Matching Network to perform Question Answering. Their model included a gated matching layer to match the question and the context, and a self-matching layer to aggregate information from the whole passage. This was done by using an RNN to summarize a concatenation of: the context embeddings at time t $(u_t)$, and a special context vector $c_t$, which was derived from an attention over the query embeddings using the previous output of the summarizer RNN and $u_t$. Thus, they generated a summary of the context and the query. The self-matching layer allowed each context embedding to gain information from its surrounding context.

\citet{DBLP:journals/corr/XiongZS16} proposed     Dynamic Coattention Network, which uses a Coattention encoder to model the interaction between the query and context. The coattention encoder involved computation of an attention matrix as the dot products between the query and context words: $L = C^TQ$. This matrix was then normalized row-wise to produce $A^Q$ (attention weights over the context for each query word), and column-wise to produce $A^C$ (attention weights over the query for each context word). $A^Q$ was further used to compute summaries of the context: $C^Q = CA^Q$. Then, $A^C$ was used to summarise the concatenation of $C^Q$ and $Q$, the summary of context words for each word of the query and the query word itself: $C^C = [Q;C^Q]A^C$. As a final step, they passed the concatenation of the context and the query aware representation of the context through a BiLSTM.

\subsection{Negation Scope Resolution}
\citet{khandelwal-sawant:2020:LREC} provide an extensive summary of papers using non-transformer based models addressing Negation Scope Resolution. We summarise their paper, and the follow-up paper by \citet{britto2020resolving}, which use transformer-based architectures to address this task. They report state-of-the-art results, and we use their models as the baseline.




\citet{khandelwal-sawant:2020:LREC} proposed using BERT \cite{bert_devlin} to address negation scope resolution. The transfer learning capability of such models allowed these models to perform much better than all previous models. They encoded the cue words in the sentence using a preprocessing strategy, by adding special tokens representing the type of the cue word before the cue words. For example:\\  \emph{I do not know the answer. $\rightarrow$ I do [cue\_tok] not know the answer.}\\
This yielded the best results to date on all datasets (BioScope Full Papers and Abstract Subcorpora, SFU Review Corpus and Sherlock Dataset).

\citet{britto2020resolving} went a step further and used XLNet and RoBERTa in place of BERT, improving results even further. 

The strategy adopted by these papers could be summarized as follows:
\begin{center}
    $ 
    Transformer \in \{BERT/XLNet/RoBERTa\}$\\
    $X = Transformer(Input Sentence)$\\ 
$Y_i = W * X_i + b ... \forall X_i \in X $
\end{center}

\subsection{Self-Attention}
Self-attention was introduced in Attention is All you Need\cite{Vaswani2017AttentionIA} as a key part of the transformer architecture. It was designed to get contextual representation of each token. The process was as follows: extracting Key $(K^{SA})$, Query $(Q^{SA})$ and Value $(V^{SA})$ vectors from each input vector, and then performing a Scaled Dot-Product Attention operation to compute the attention weights for each token $i$, using the token's query vector $Q^{SA}_i$ and all Key vectors $K^{SA}$. These weights were used to summarise the Value vectors $V^{SA}$, to compute the contextual representation.
This operation was repeated multiple times in parallel (using different weight vectors), and their outputs combined using a weight matrix. This was called Multiheaded Self Attention. We can summarise the above process using the following equations:
$$ Z_i^{k} = \sum_j softmax(\frac{Q^{SA}_i \bullet K^{SA}_j}{\sqrt{d_k}})V^{SA}_j $$

$$ Z = Concat_k(Z^k)(W^O)^T $$

\section{Orthogonal Attention}
Inspired by self-attention \cite{Vaswani2017AttentionIA}, we look to use the key-query-value structure and the multiheaded process to create an Attention mechanism to address Cloze-Style tasks. We first detail the overall modelling framework, and then explain individual components. 

We use the following notational convention: $Q$ is Query Inputs, $C$ is Context Input, $Q^S$ is Query Vectors for Orthogonal Attention, $K^S$ is Key Vectors for Orthogonal Attention, $V^S$ is Value Vectors for Orthogonal Attention, $Q^D$ is Query Vectors for Dot-Product Attention and $C^D$ is Context Vectors for Dot-Product Attention.

\subsection{Orthogonal Attention Encoder Block (OA)}

\begin{figure}[!h]
    \centering
    \includegraphics[width=0.55\linewidth]{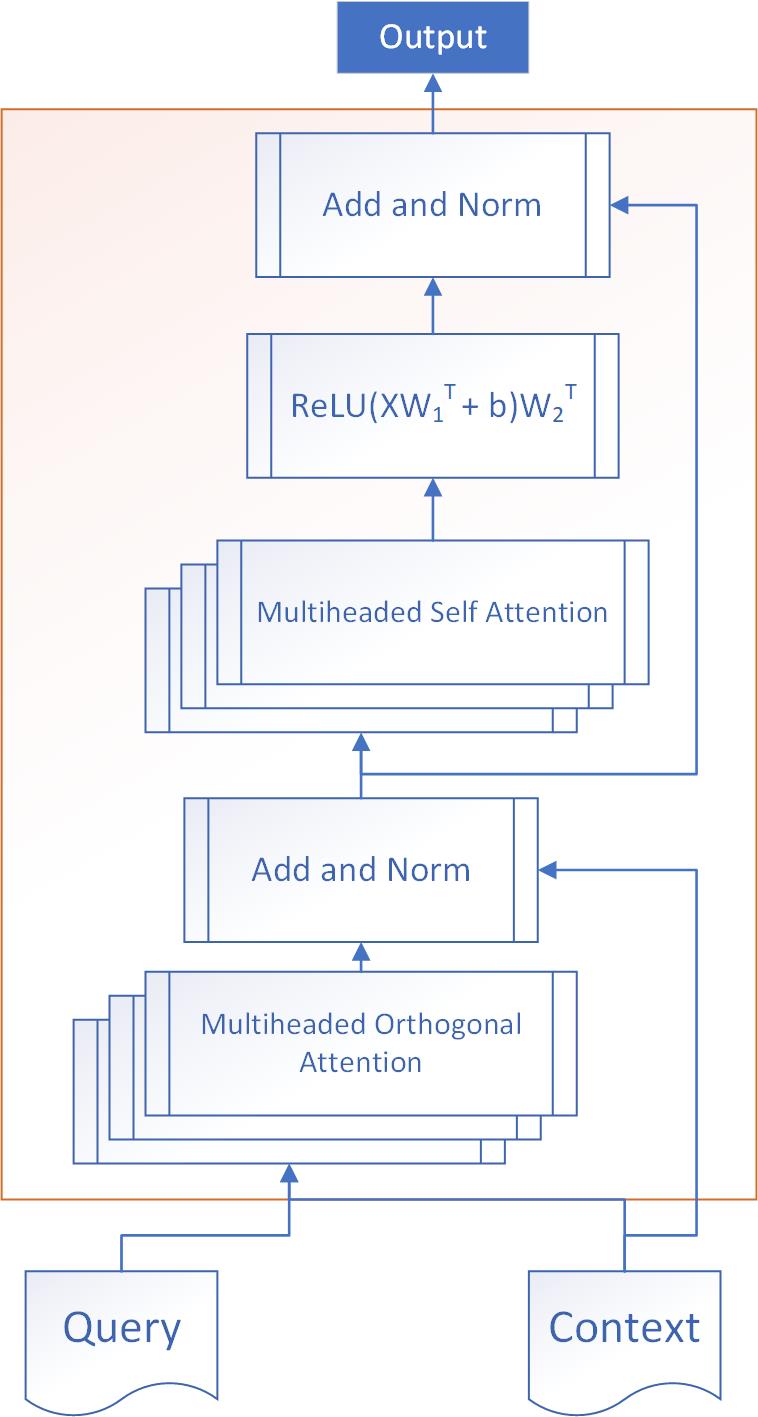}
    \caption{Orthogonal Attention Block}
    \label{fig:oa}
\end{figure}
\begin{center}
    \textbf{Multiheaded Orthogonal Attention:} $Z = f(C,Q)$\\
    \textbf{Residual Connection:} $X_1 = Z + C$ \\
    \textbf{Layer Normalization:} $X_2 = LayerNorm(X_1)$  \\
    \textbf{Multiheaded Self-Attention:} $X_3 = SelfAtt(X_2)$  \\
    \textbf{2 Fully Connected Layers:} $X_4 = ReLU(X_3W^T + b)^TW_2$ \\
    \textbf{Residual Connection:} $X_5 = X_4 + X_2$ \\
    \textbf{Layer Normalization:} $C' = LayerNorm(X_5)$
\end{center}  
This is quite similar to the transformer encoder block proposed in \citet{Vaswani2017AttentionIA}.

We use a Multiheaded Self-Attention module after the Multiheaded Orthogonal Attention module so as to allow the query-aware contextual representations to exchange information among themselves. This is similar in function to the self-matching layer by \cite{wang-etal}.

\subsection{Multiheaded Orthogonal Attention}
First, we detail the design of an individual head of Orthogonal Attention:

\begin{figure}[!htb]
    \centering
    \includegraphics[width=\linewidth]{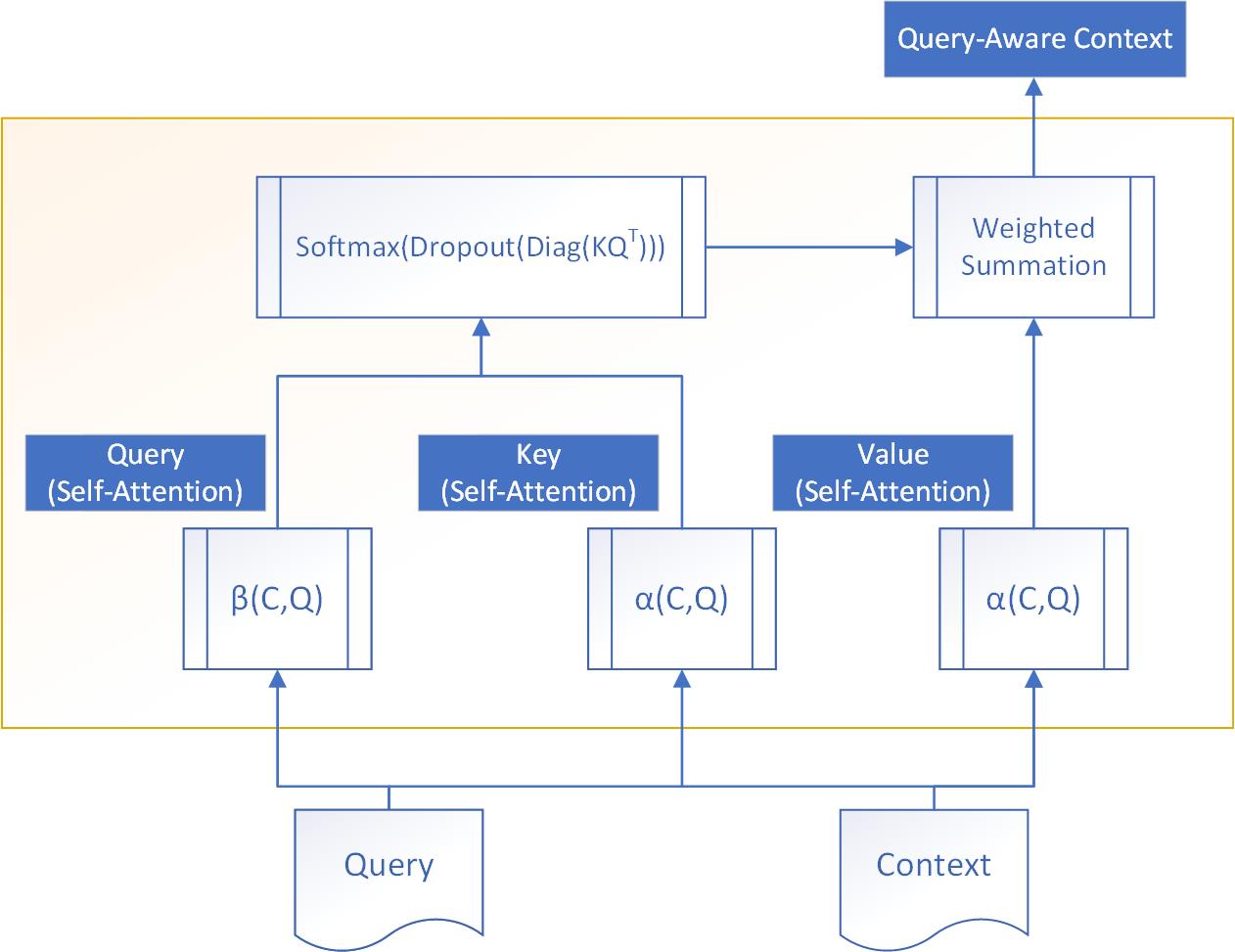}
    \caption{Orthogonal Attention Single Head}
    \label{fig:oa-single-head}
\end{figure}

We use a function $\alpha(C,Q)$ to generate $\|q\|*\|c\|$ key-value pairs ($K^S_{ij}$ and $V^S_{ij}$), and $\beta(C,Q)$ to generate $\|q\|$ query vectors $(Q^S_j)$. Then, we summarise the value vectors using the key and query vectors, similar to self-attention. Specifically, the summation is performed for each context word ($i's$) over all $j's$ in $V^S_{ij}$. This generates a context word representation that is query-aware. This operation is performed in a multiheaded fashion. Specifically, the following equations can be used to summarise the entire process:
$$C^Q_i = \sum_j softmax(\frac{K^S_{ij} \bullet Q^S_j}{\sqrt{d_k}})V^S_{ij}$$

Specifically,
\begin{center}
    \textbf{Query:} $Q$ $... [n,d]$\\
    \textbf{Context:} $C$ $... [m,d]$ \\
    $Q^S = \beta(Q,C)$ $... [n,d_k] $ \\
    $K^S = \alpha(Q,C)$ $...[m,n,d_k]$ \\
    $V^S = \alpha(Q,C)$ $...[m,n,d_v]$ \\
    \textbf{Attention Weights:}\\ $W^S = Dropout(Softmax(Diag(K^S(Q^S)^T)))$ $...[m,n,1]$ \\
    $C^Q_i = Sum(W^S * V^S, axis=1).squeeze(1)$ $...[m,d_v]$ \\
\end{center}

$C^Q_i$ represent the output of the $i^{th}$ Orthogonal Attention head. $C^O$ represents the output of the Multiheaded Orthogonal Attention Layer 
\begin{center}
    $C^O = f(C,Q) =$ $Concat_k([C^Q_k], axis=1)(W^D)^T$ $...[m,d]$
\end{center}
Like Self-Attention, we choose $dk = dv = \frac{d}{n\_heads}$. $W^D$ is used to project the output back to a $d$ dimensional space.


\subsection{Orthogonal Attention Variants}
Orthogonal Attention models the interaction between the Query and Context using 2 functions, $\alpha$ and $\beta$, which are used to produce the key, query and value vectors $K^S, Q^S$, and $V^S$. In this section, we detail 4 choices for $\alpha$ and $\beta$. 

We experiment with 2 primary modes of interaction between the query vectors $Q_j$ and context vectors $C_i$ to generate the $K^S$ and $V^S$: 
\begin{itemize}
    \item An elementwise multiplication operation (this is used in OA-EM and OA-EMB). This approach is similar to \cite{DBLP:journals/corr/DhingraLCS16}, who also relied on element-wise multiplicative interaction between query and context words. This could be thought of as the query acting as a filter for each context dimension.
    \item A 1D Convolutional operation using Query vectors $Q_j$ as filters to convolve over the Context vectors $C_i$ (this is used in OA-C and OA-CA). The intuition is that instead of an elementwise multiplicative filter, we generate filters from each query word that could be convolved over the context word embeddings.
\end{itemize}  

To generate $Q^S$, we experiment with 2 choices:
\begin{itemize}
    \item Only using the Query input $Q$ (this is used in OA-C and OA-EM).
    \item Using both the Query $Q$ and the Context $C$ (this is used in OA-CA and OA-EMB). This choice is inspired by \citet{DBLP:journals/corr/SeoKFH16}, who proposed that using bidirectionality helps the model.
\end{itemize}

\subsubsection{OA-EM}
Here, $\alpha^{EM}$ uses Elementwise Multiplication to model the interaction between each context word-query word pair. $\beta^{EM}$ only uses $Q$ to generate $Q^S$. Formally,

\begin{figure}[!htb]
    \centering
    \includegraphics[width=\linewidth]{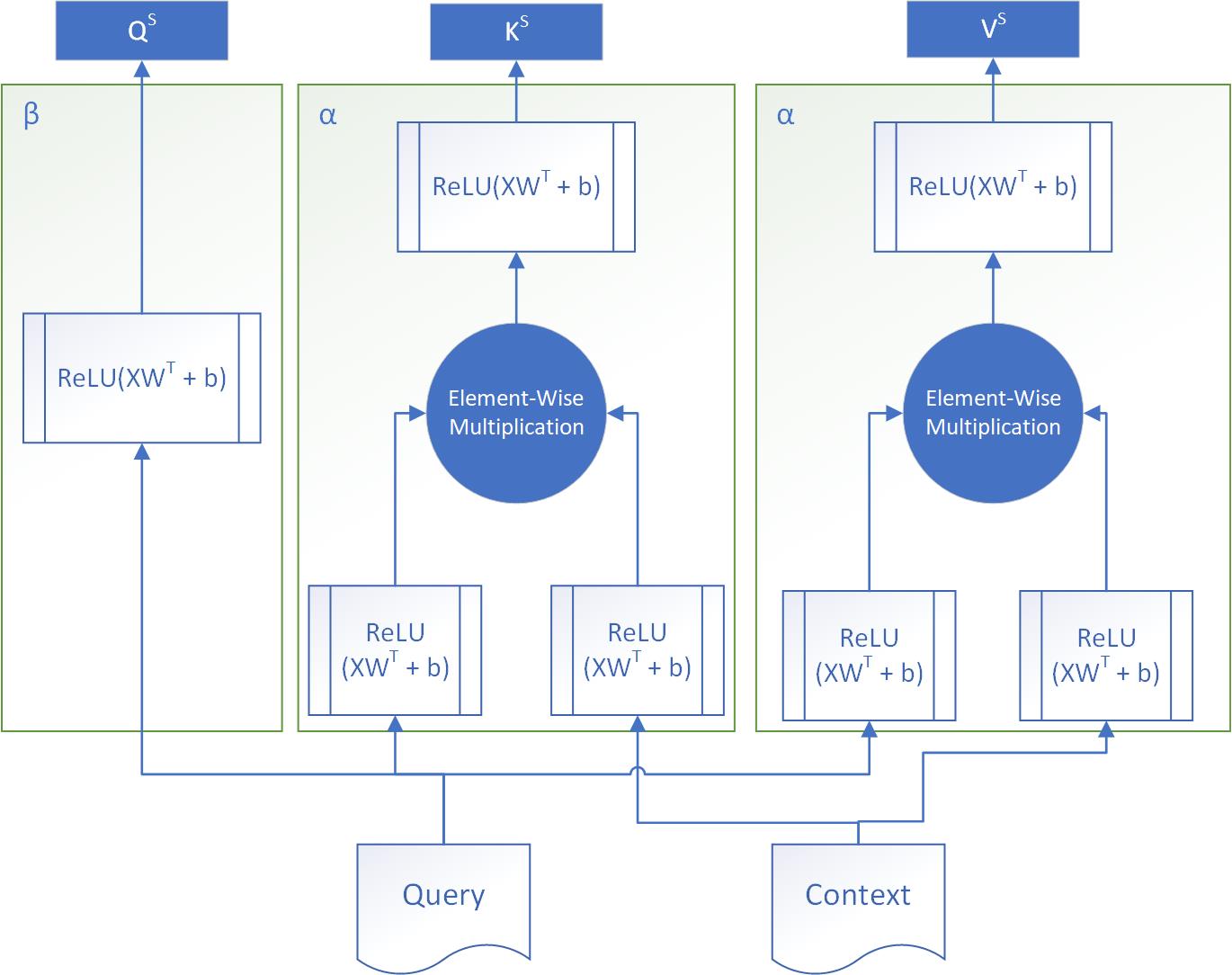}
    \caption{OA-EM: Internal Design}
    \label{fig:oa-em}
\end{figure}
\begin{itemize}
\setlength\itemsep{0em}
    \item $\alpha^{EM}(C,Q)$:
    \begin{center}
    \setlength\itemsep{0em}
        \textbf{Linear:} $C^1 = ReLU(CW_0^T + b_0$) $...[m,d_k]$ \\
        \textbf{Linear:} $Q^1 = ReLU(QW_1^T + b_1$ $...[n,d_k]$ \\
        \textbf{Elementwise Multiplication:} $X^1 = C^1.reshape(n,1,d_k) \odot Q^1.reshape(1,n,d_k)$ $...[m,n,d_k]$ \\
        \textbf{Linear:} $\alpha^{EM}(C,Q) = ReLU(X^1W_2^T + b_2)$ $...[m,n,d_k]$
    \end{center}
    \item $\beta^{EM}(C,Q):$
    \begin{center}
        \textbf{Linear:} $\beta^{EM}(C,Q) = ReLU(QW_3^T + b_3)$ $...[n,d_k]$
    \end{center}
\end{itemize}
  
\subsubsection{OA-EMB}
Here, $\alpha^{EMB}$ uses Elementwise Multiplication to model the interaction between each context word-query word pair, just like in $\alpha^{EM}$. Unlike OA-EM though, $\beta^{EMB}$ generates $Q^S$ as context-aware query representations using elementwise multiplication. Specifically, it uses the dot product attention as implemented in the PyTorchNLP Library \footnote{https://pytorchnlp.readthedocs.io} to calculate a summary of the context $C$ per query word $Q_j$. This generates a vector summary of $C$ for each query word, which is then multiplied elementwise with the query vectors $Q$ themselves to get the context-aware query representation $Q^S$. Formally,

\begin{figure}[!htb]
    \centering
    \includegraphics[width=\linewidth]{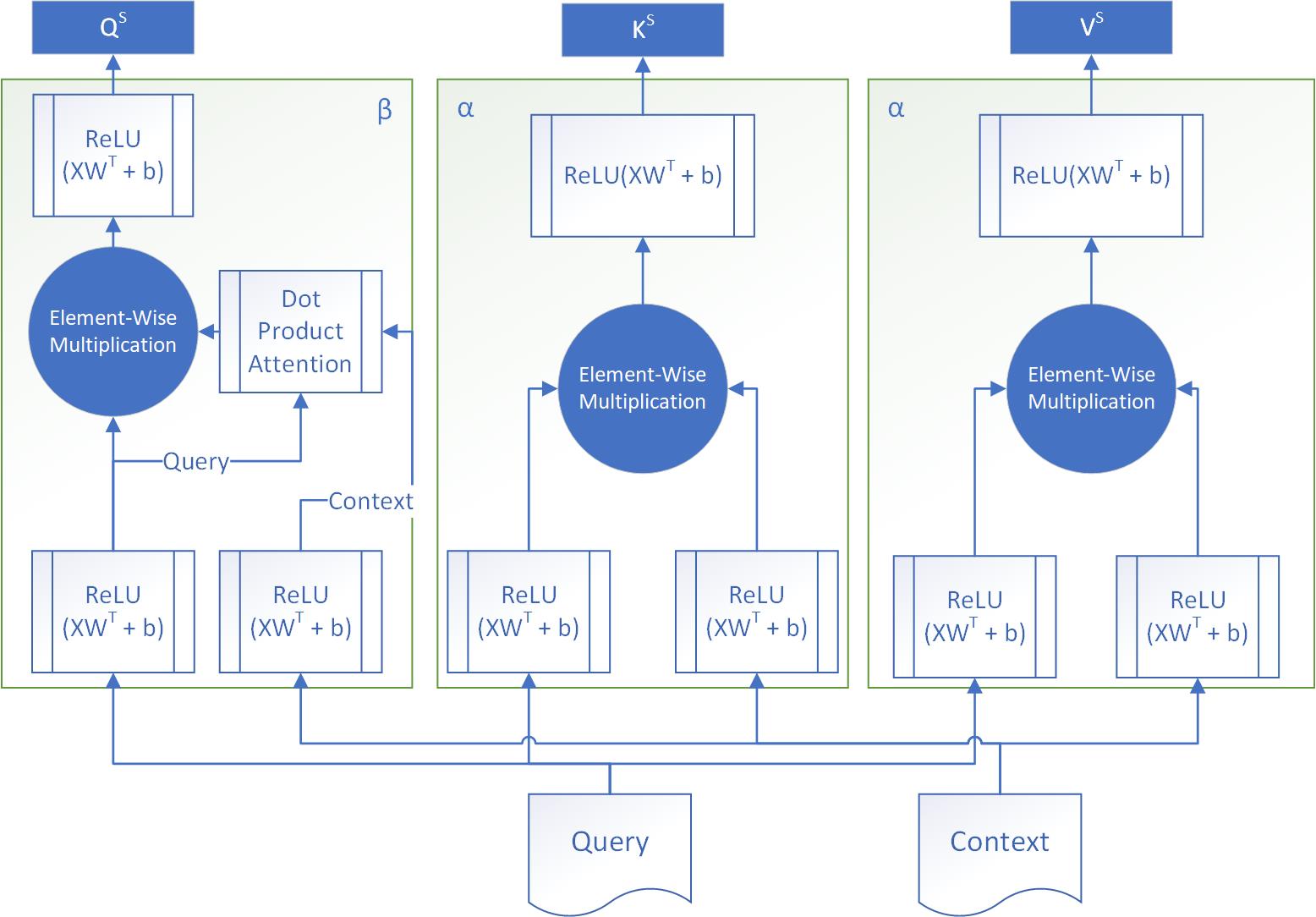}
    \caption{OA-EMB: Internal Design}
    \label{fig:oa-emb}
\end{figure}
\begin{itemize}
    \item $\alpha^{EMB}(C,Q)$:
    \begin{center}
    \textbf{Elementwise Multiplication}: $\alpha^{EMB}(C,Q) = \alpha^{EM}(C,Q)$ $...[m,n,d_k]$
    \end{center}
    \item $\beta^{EMB}(C,Q):$
    \begin{center}
        \textbf{Linear:} $C^1 = ReLU(CW_0^T + b_0)$ $...[m,d_k]$ \\
        \textbf{Linear:} $Q^1 = ReLU(QW_1^T + b_1)$ $...[n,d_k]$ \\
        \textbf{Dot Product Attention:} $C^Q = Attention(Q^D = Q^1, C^D = C^1)$ $...[n,d_k]$ \\
        \textbf{Elementwise Multiplication:} $Q^2 = Q^1 \odot C^Q$ $...[n,d_k]$\\
        \textbf{Linear:} $\beta^{EMB}(C,Q) = ReLU(Q^2W_2^T + b_2)$ $... [n,d_k]$
    \end{center}
\end{itemize}

\subsubsection{OA-C}
Here, $\alpha^{C}$ uses 1D Convolutional Operation to model the interaction between each context word-query word pair. Specifically, we use $Q$ to generate 1D Convolutional filters, which are then convolved over each context word $C_i$ separately. We use $\sqrt{d_k}$ filters of size $\sqrt{d_k}$, with a stride of $\sqrt{d_k}$. The resulting feature maps are flattened to produce embeddings for each context word-query word pair, from which $K^S$ and $V^S$ are generated. $\beta^{C}$ only uses $Q$ to generate $Q^S$, just like $\beta^{EM}$. Formally,

\begin{figure}[!htb]
    \centering
    \includegraphics[width=\linewidth]{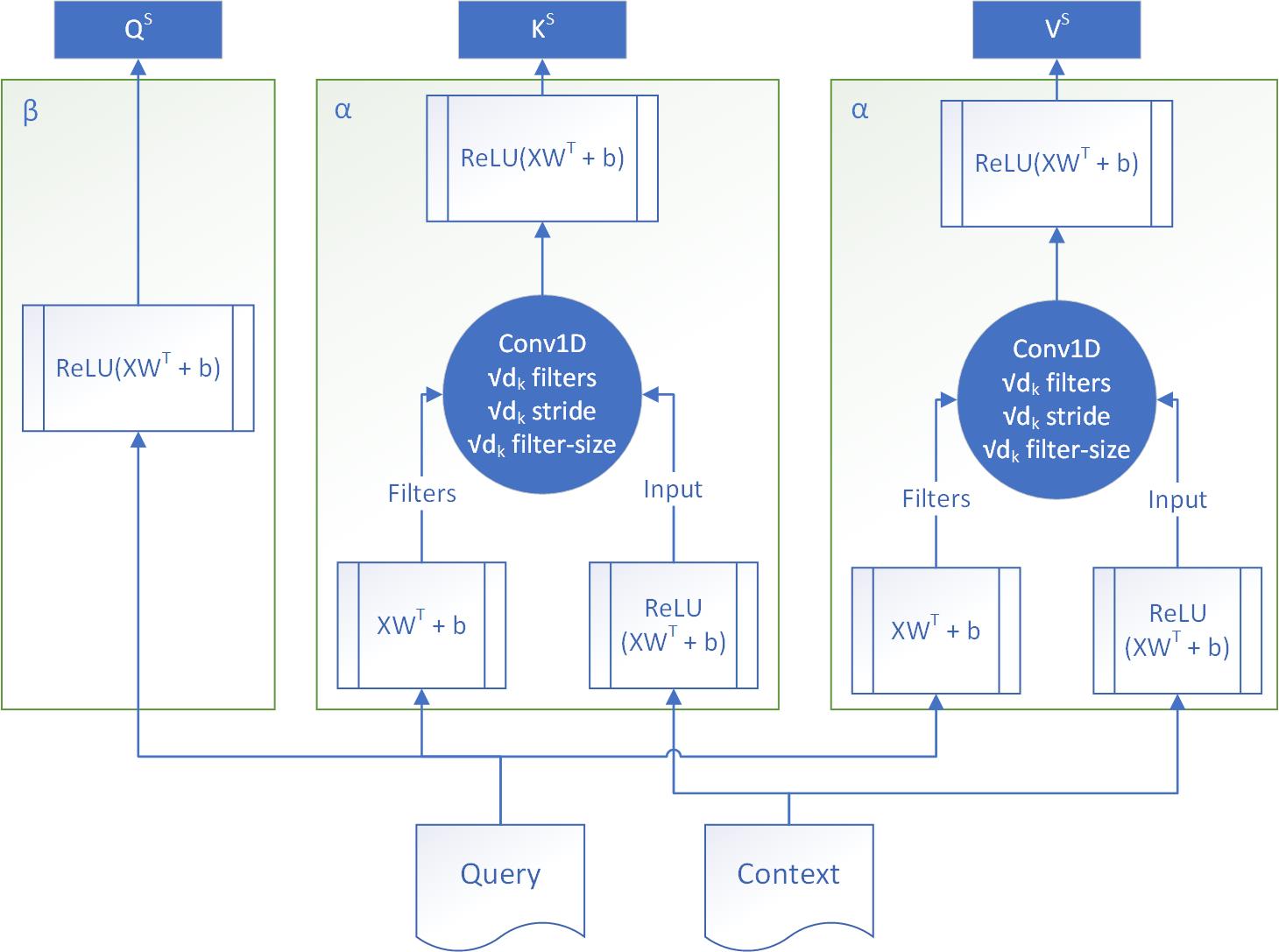}
    \caption{OA-C: Internal Design}
    \label{fig:oa-c}
\end{figure}
\begin{itemize}
    \item $\alpha^C(C,Q):$\\
    \begin{center}
        \textbf{Linear:} $C^1 = ReLU(CW_0^T + b_0)$ $...[m,d_k]$ \\
        \textbf{Linear (Generate convolutional filters):} $W_{conv} = QW_1^T + b_1$ $...[n,\sqrt{d_k}]$\\
        \textbf{Linear (Generate convolutional biases):} $b_{conv} = QW_2^T + b_2$ $...[n,1]$\\
        \textbf{1D Convolution:} $X = Conv1D(input = C^1, filters=(W_{conv},b_{conv}),stride=\sqrt{d_k})$ $...[m,n,d_k]$\\
        \textbf{Linear:} $\alpha^C(C,Q)=ReLU(XW_3^T+b_3)$ $...[m,n,d_k]$
    \end{center}
    \item $\beta^C(C,Q):$
    \begin{center}
        \textbf{Linear:} $\beta^C(C,Q) = \beta^{EM}(C,Q)$ $...[n,d_k]$
    \end{center}
\end{itemize}
\subsubsection{OA-CA}
Here, $\alpha^{CA}$ uses 1D Convolutions to model the interaction between each context word-query word pair, just like in $\alpha^{C}$. Unlike OA-C though, $\beta^{CA}$ uses $C$ and $Q$ to generate $Q^S$. Specifically, we perform a dot product attention mechanism over $C$ using each individual query word $Q_j$ to generate a vector summary of $C$ for each query word (like in OA-EMB), which are then used to generate filters for a 1D Convolution over the corresponding query word  (from which the filter was generated). We use $\sqrt{d_k}$ filters of size $\sqrt{d_k}$, with a stride of $\sqrt{d_k}$. The resulting feature maps are flattened to produce embeddings for each query word, from which $Q^S$ are generated. Formally,

\begin{figure}[!htb]
    \centering
    \includegraphics[width=\linewidth]{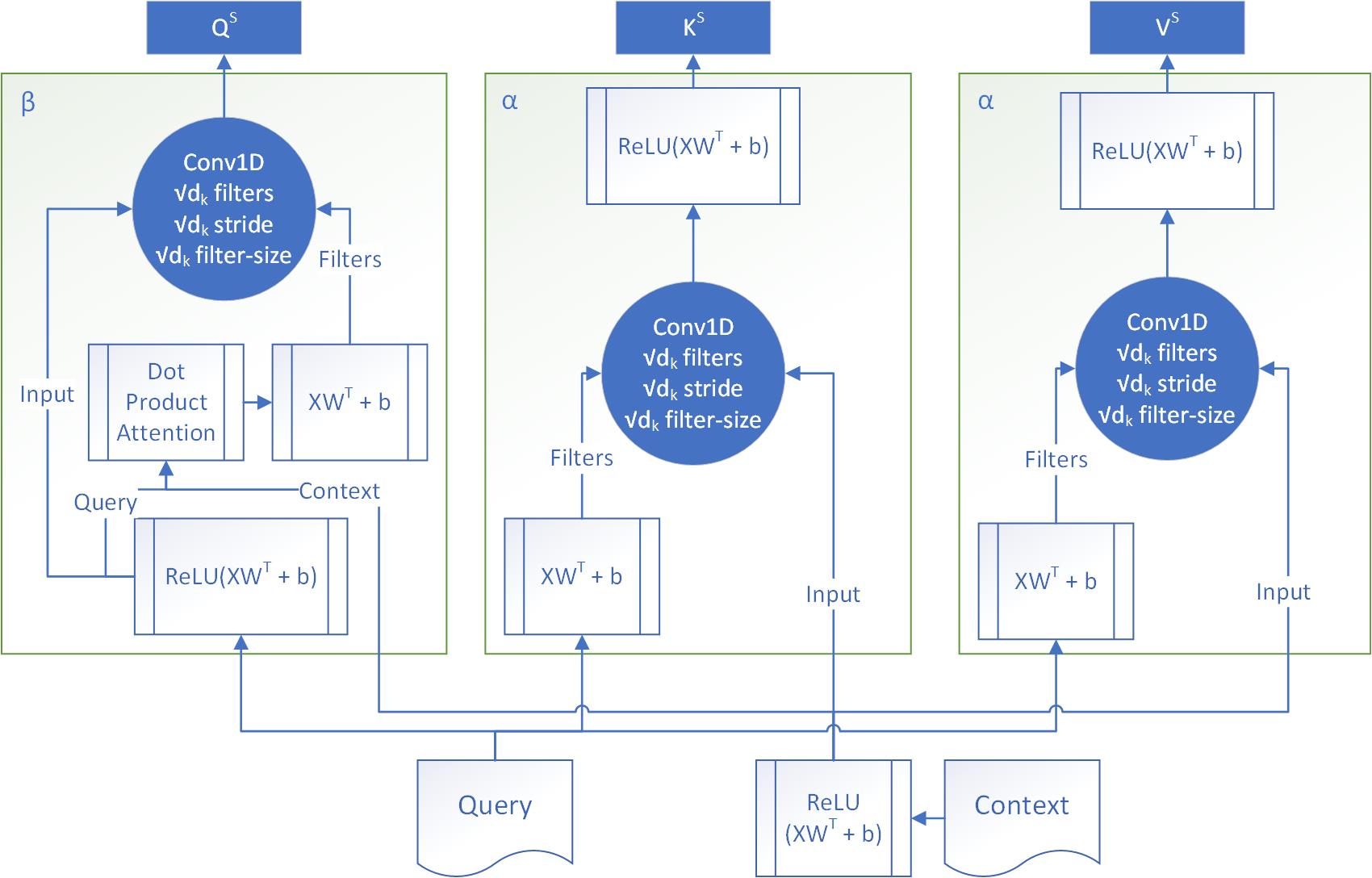}
    \caption{OA-CA: Internal Design}
    \label{fig:oa-ca}
\end{figure}
\begin{itemize}
    \item $\alpha^{CA}(C,Q)$:
    \begin{center}
        \textbf{Conv1D:} $\alpha^{CA}(C,Q)=\alpha^{C}(C,Q)$ $...[m,n,d_k]$
    \end{center}
    \item $\beta^{CA}(C,Q):$
    \begin{center}
        \textbf{Linear:} $C^1 = ReLU(CW_0^T + b_0)$ $...[m,d_k]$ \\
        \textbf{Linear:} $Q^1 = ReLU(QW_4^T + b_4)$ $...[n,d_k]$\\
        \textbf{Dot Product Attention:} $C^Q = Attention(Q^D = Q^1, C^D = C^1)$ $...[n,d_k]$\\
        \textbf{Linear (Generate convolutional filters):} $W_{conv} = C^QW_5^T + b_5$ $...[n,\sqrt{d_k}]$\\
        \textbf{Linear (Generate convolutional biases):} $b_{conv} = C^QW_6^T + b_6$ $...[n,1]$\\
        \textbf{1D Convolution:} $\beta^{CA}(C,Q) =  Conv1D(input = Q^1, filters=(W_{conv},b_{conv}),stride=\sqrt{d_k})$ $...[n,d_k]$\\
    \end{center}
\end{itemize}

\section{Experimentation Details}

The model architecture is a straightforward modification of \citet{khandelwal-sawant:2020:LREC}, by passing the output of the XLNet-base model to 2 Orthogonal Attention Encoder blocks. Finally, we add a linear layer to get the output probabilities. We also use Dropout and Residual Connections to regularize the model and stabilize training.

The model architecture can be summarised as follows:
\begin{center}
\textbf{Sentence Embeddings:} $X_1 = XLNet-base(X)$ \\
\textbf{Dropout:} $X_2 = Dropout(X_1)$ \\
\textbf{Orthogonal Attention Encoder:} $X_3 = OA(X_2, X_2[cue\_ids])$\\
\textbf{Orthogonal Attention Encoder:} $X_4 = OA(X_3, X_3[cue\_ids])$\\
\textbf{Dropout:} $X_5 = Dropout(X_4)$ \\
\textbf{Residual Connection:} $X_6 = X_5 + X_1$ \\
\textbf{Dropout:} $X_7 = Dropout(X_6)$ \\
\textbf{Linear:} $Y = X_7W^T + b$
\end{center}
We experiment with 4 datasets:
\begin{itemize}
    \setlength\itemsep{0em}
    \item BioScope Abstracts Subcorpora (BA): 1075 samples
    \item BioScope Full Papers Subcorpora (BF): 235 samples
    \item SFU Review Corpus (SFU): 2205 samples
    \item *sem 2012 Shared Task dataset (Sherlock): 615 samples
\end{itemize}

We perform a 10-fold cross validation. When the Sherlock dataset is involved, we use the train-dev-test split provided by *sem 2012 organizers, so the 10-fold CV becomes a 10-run average over the test set provided. We report the scores for the Token level F1, which is an F1 score over the number of word level label matches. Using the proposed model architecture, we experiment with the 4 proposed variants of Orthogonal Attention. For each of the variants, we also experiment with 2 preprocessing techniques for the input sentence:
\begin{itemize}
    \setlength\itemsep{0em}
    \item \textbf{Normal:} The input sentence is passed as is. We hypothesise that this approach will require the model to learn the dependencies between the cue words and their scope via leveraging the multiheaded orthogonal attention blocks, as the XLNet layer does not know which words are the cue words in the sentence.
    \item \textbf{Augment:} Similar to the Augment method used in prior work (\cite{khandelwal-sawant:2020:LREC}, \cite{britto2020resolving}), the cue words are preceded with a special token (tok[0]). This way, we explicitly tell the XLNet layer about the cue words as well, and thus the Orthogonal Attention Encoder layers would be able to enhance the representation of each token, making it easier for the linear layer to find the scope of the cue.
\end{itemize}

While the Normal Preprocessing strategy would be the default method for a Cloze-style task, we observed that our model overfit to the training set, due to the limited amount of training data (ranging from 200 samples to 2200 samples). Each Orthogonal Attention layer had around 4-7 million parameters which had to be learnt from scratch (as shown in Table \ref{fig:num_params}). To reduce overfitting, we experimented with also allowing access to cue information to the XLNet layer,so that it generated meaningful cue-aware embeddings, via the Augment preprocessing technique. In our analysis, we show that all the Orthogonal Attention variants give us a gain in performance over using XLNet-base only.

We use a differential learning rate strategy, where XLNet-base is finetuned using an initial learning rate of 3e-5 and the Orthogonal Attention Encoder Layers are trained using an initial learning rate of 1e-4, both using the Adam optimizer.

We use a batch size of 32, a dropout probability of 0.3, and the maximum input sequence length as 128. We train for 60 epochs, with an early stopping on the validation set F1 with a patience of 6 epochs. From k folds that the model is divided into, one fold is kept as the test set while a validation set (for early stopping) is sampled from the k-1 folds, whose size is the same as the test set.

We also perform a 10-fold Cross-Validation for the model proposed by  \citet{britto2020resolving}, using XLNet as the base model (which is the state-of-the-art model for BF, BA and SFU), to accurately compare the results. This model is referred to as Baseline in all further results.

\section{Results and Analysis}

\begin{table}[!htb]
    \centering
    \includegraphics[width=0.8\linewidth]{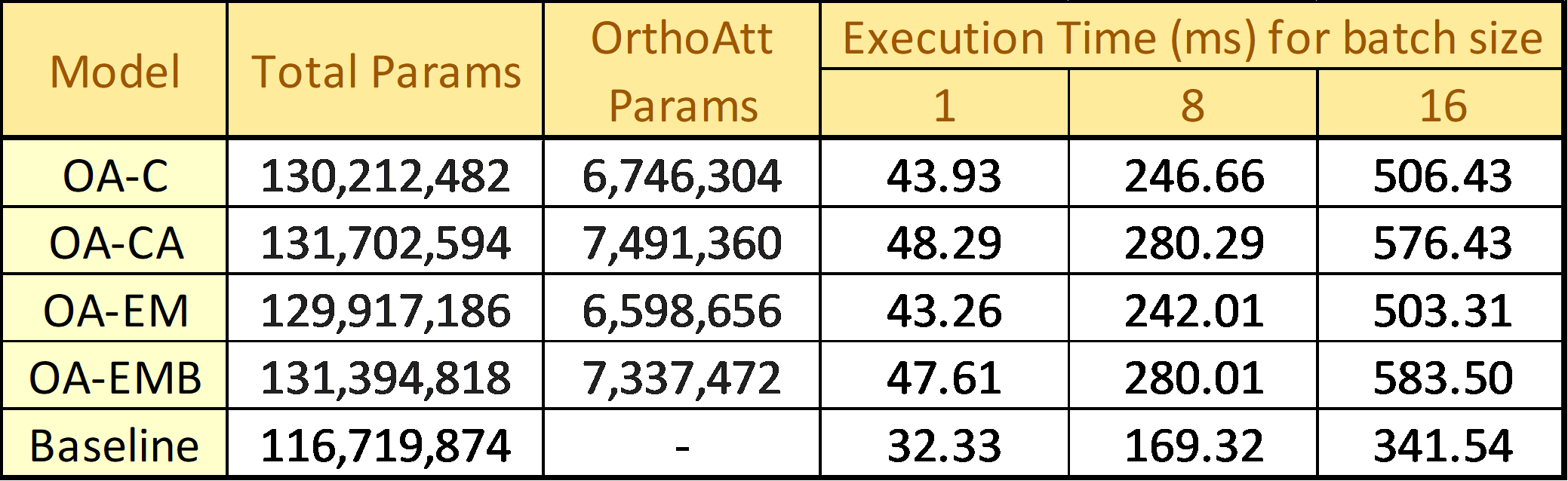}
    \caption{Number of Model Parameters}
    \label{fig:num_params}
\end{table}
Table \ref{fig:num_params} contains a summary of the model sizes and the execution times for an inference run with various batch sizes. We see that OA-C and OA-EM have similar execution profiles, and that OA-CA and OA-EMB (the bidirectional variants) have similar execution profiles.
\begin{table}[!htb]
    \centering
    \includegraphics[width = 0.95 \linewidth]{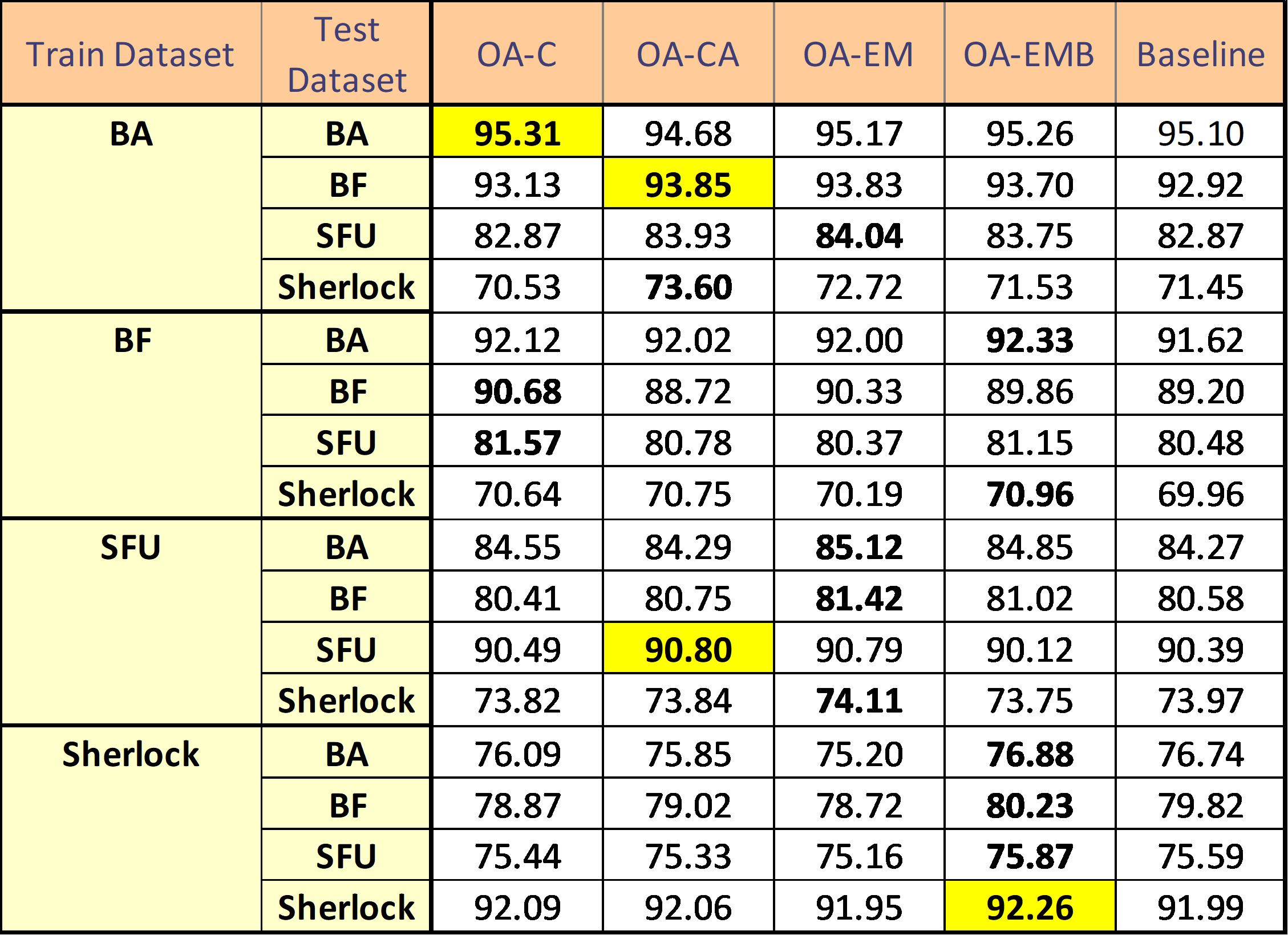}
    \caption{Results for Augment Preprocessing Method}
    \label{tab:augment}
\end{table}
\begin{table}[!htb]
    \centering
    \includegraphics[width = 0.8\linewidth]{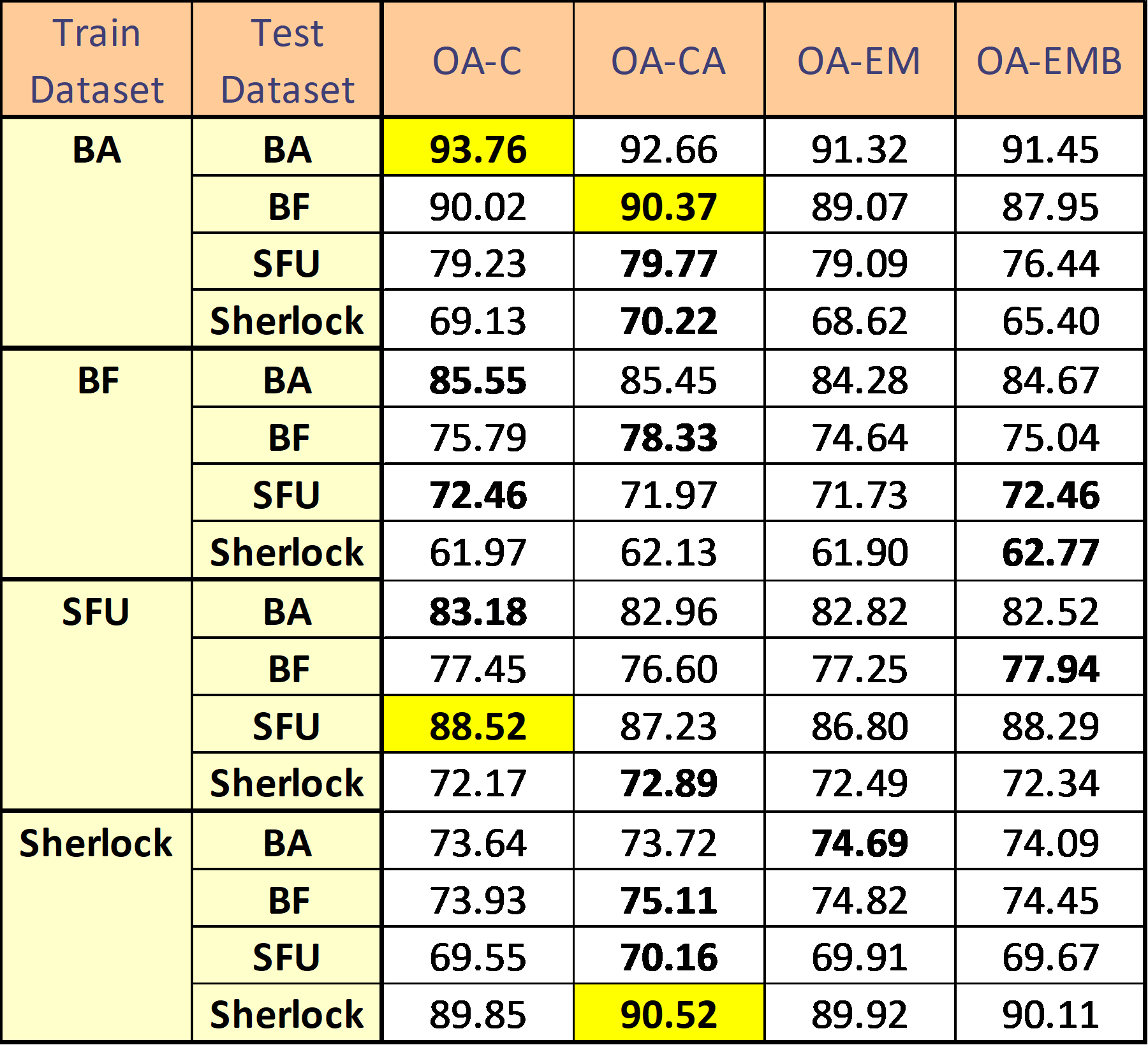}
    \caption{Results for Normal Preprocessing Method}
    \label{tab:normal}
\end{table}

Tables \ref{tab:augment} and \ref{tab:normal} contain the results for Augment and Normal Preprocessing method respectively (Token-level F1 score (Macro Average over a 10-fold CV)). Across a row, the best model for that train dataset-test dataset combination is highlighted in bold. The best score for a given test dataset is highlighted in yellow. We observe that:
\begin{itemize}
    \setlength \itemsep{0em}
    \item
    \begin{table}[!htb]
    \centering
    \includegraphics[width =  0.75\linewidth]{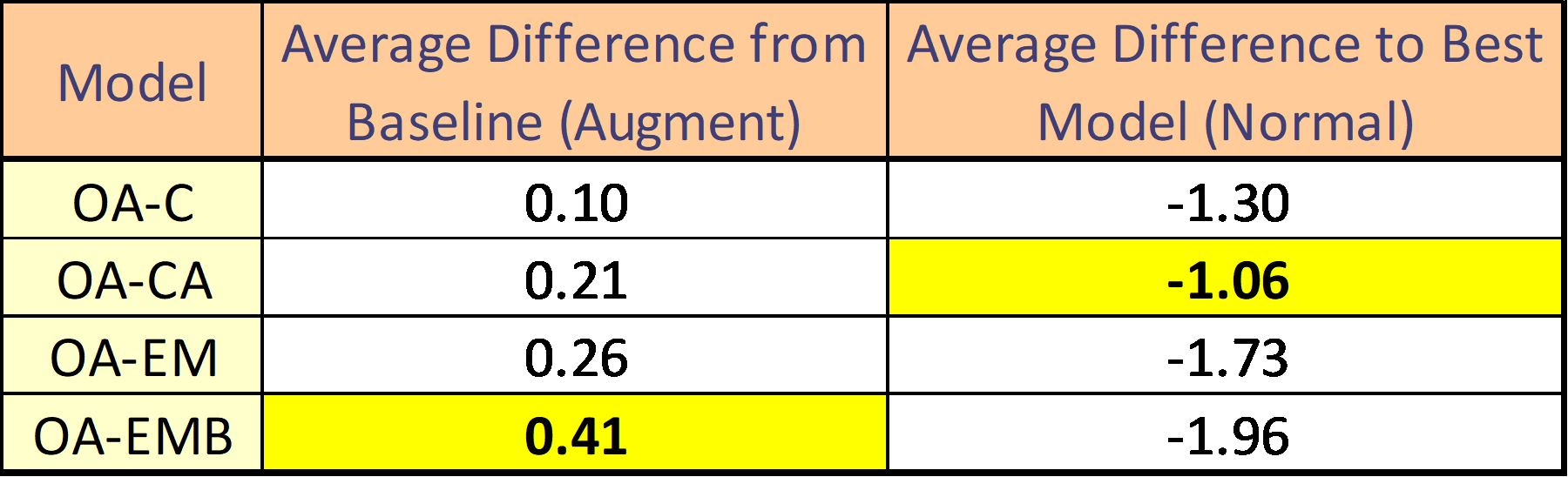}
    \caption{Performance Difference (F1 points)}
    \label{tab:diff_to_baseline}
    \end{table}
    Table \ref{tab:diff_to_baseline} shows that for the Augment preprocessing method, all models yield improvements over the baseline(non-OA) for Augment preprocessing method.To compute the second column, we take the difference of the scores of that model to the best score that any model had on that train dataset-test dataset combination. It also shows that the 2 most promising variants are OA-EMB and OA-CA (the bidirectional variants).
    \item 
    \begin{figure}[!htb]
    \centering
    \includegraphics[width =  0.9\linewidth]{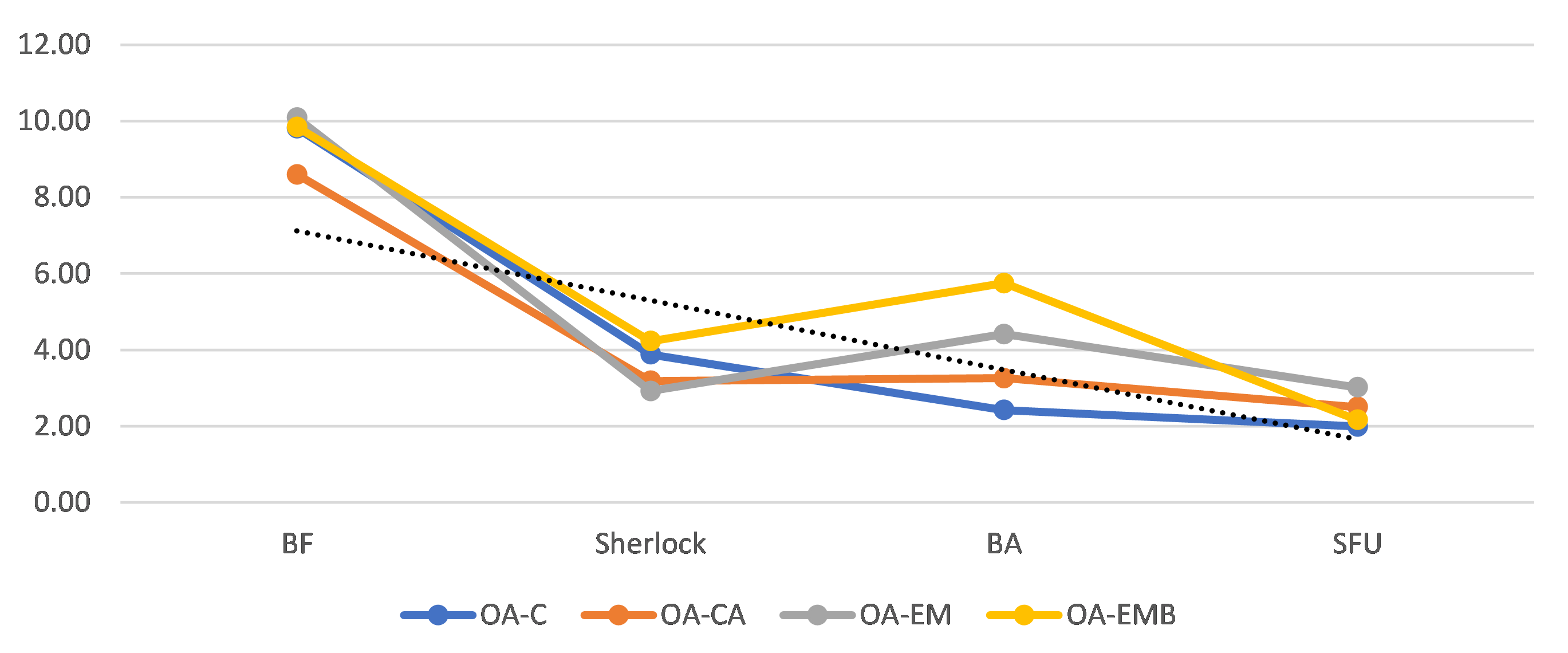}
    \caption{Performance Difference between Augment and Normal (F1 Points)}
    \label{tab:perf_diff_augment_normal_table}
    \end{figure}
    Figure \ref{tab:perf_diff_augment_normal_table} contains the difference between performance for the a model when the only difference was the preprocessing method used, averaged. We observe that the while the Normal preprocessing method has a significant difference in performance compared to Augment, but that this difference decreases with increasing dataset sizes (supporting our earlier hypothesis of overfitting).
\end{itemize}
\begin{table}[!htb]
    \centering
    \includegraphics[width=0.9\linewidth]{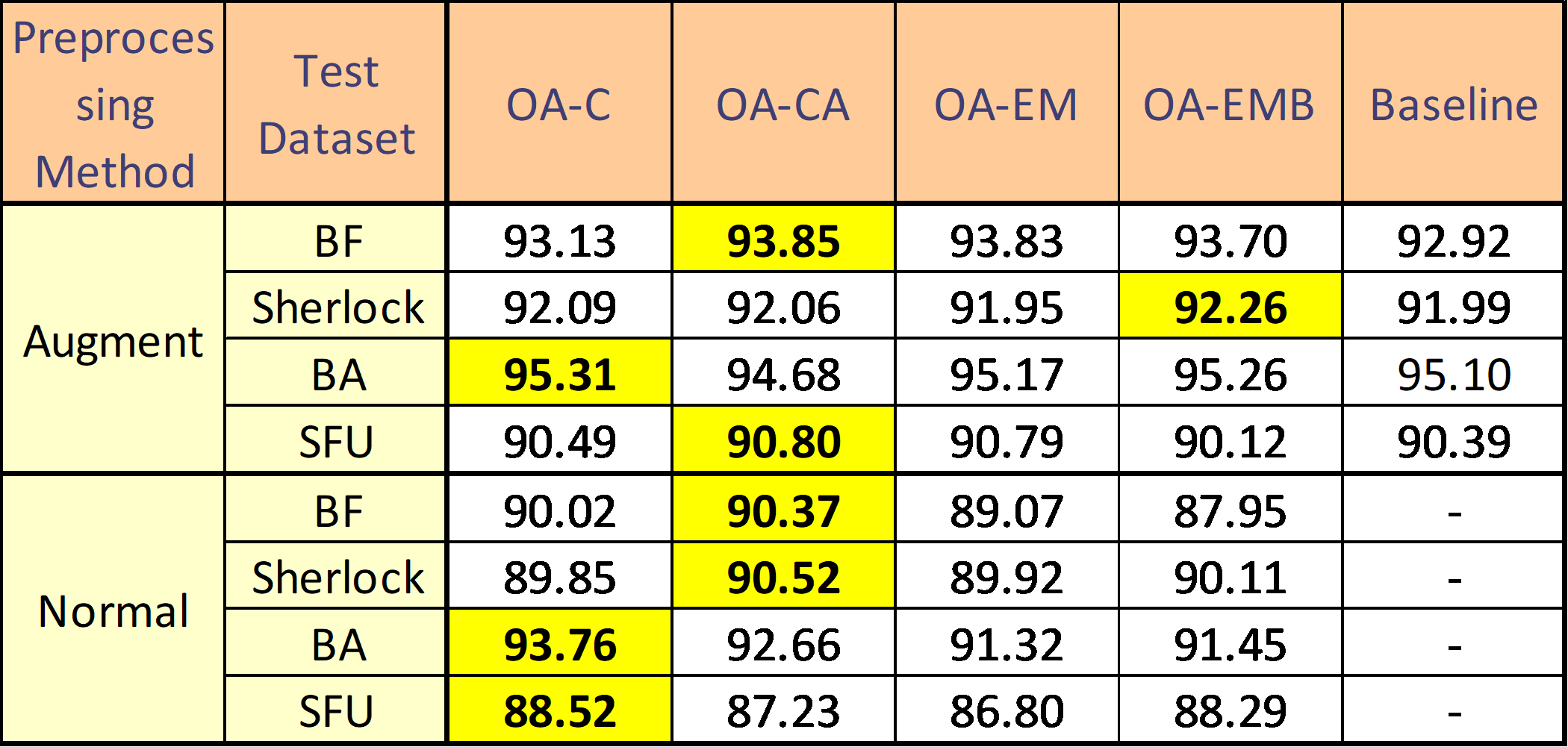}
    \caption{Summary of Results}
    \label{fig:summary}
\end{table}
A summary of the best results by each model variant is shown in Table \ref{fig:summary}. A final summary of the results in comparison to the previous state-of-the-art results is shown in Table \ref{tab:final_summary_tab}.
\begin{table}[!htb]
    \centering
    \includegraphics[width=0.95\linewidth]{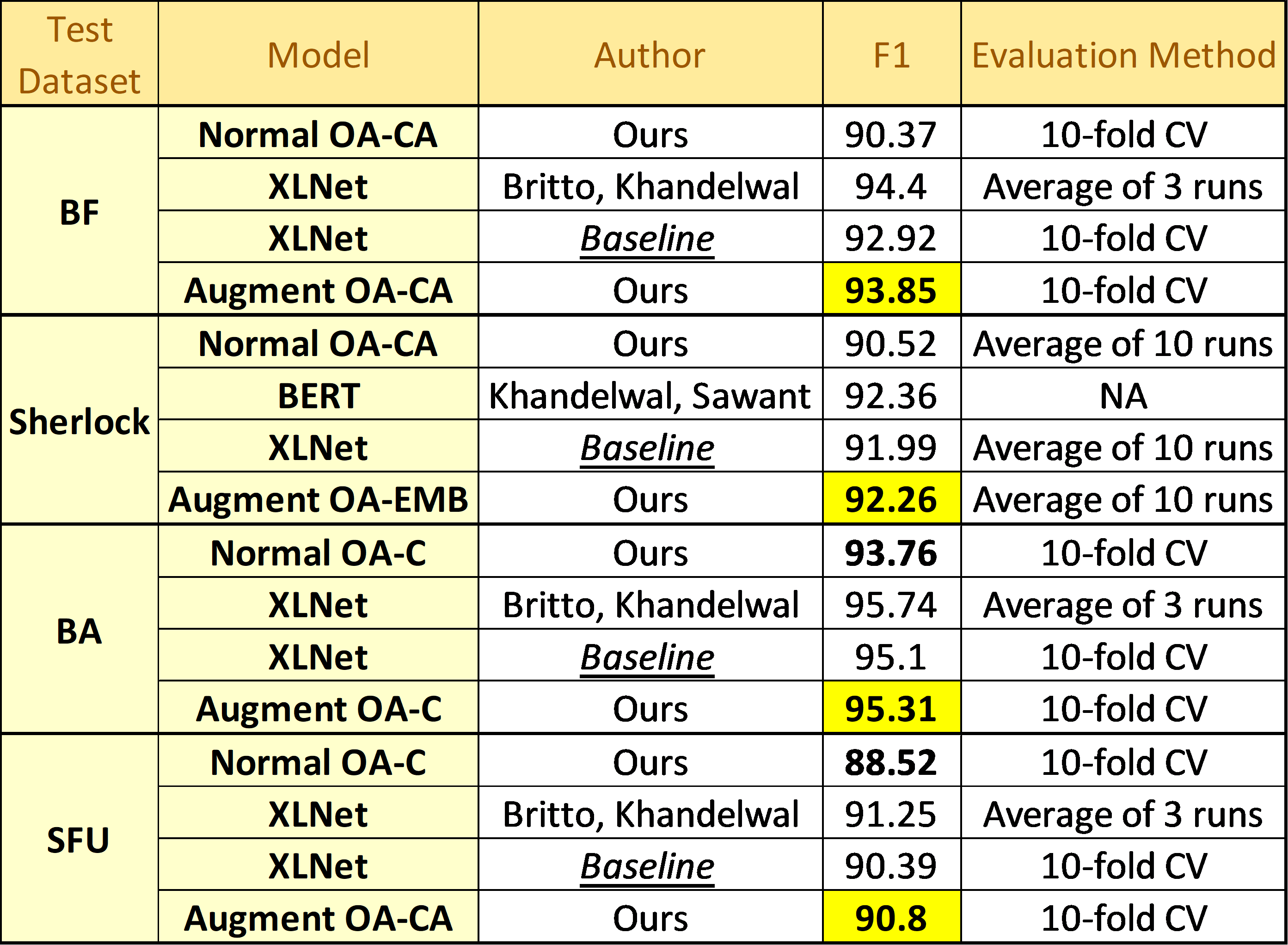}
    \caption{Final Summary Table}
    \label{tab:final_summary_tab}
\end{table}

\section{Conclusion and Future Scope}
In this paper, we proposed a novel approach to solving Negation Scope Resolution by viewing it as a Cloze-Style task, and also proposed a novel Cloze-Style Attention mechanism called Orthogonal Attention. We proposed 4 such variants. Our results showed that Orthogonal Attention is very effective as a Cloze-Style Attention mechanism, and using it with the current state-of-the-art models (XLNet-base) gives us an increase in performance over them. Thus, we report the best results till date on Negation Scope Resolution.

Future work could utilize Orthogonal Attention to address Question Answering and other Machine Reading Comprehension tasks. The Orthogonal Attention framework could be used to create better variants of Orthogonal Attention, which could better model the interaction between the Query and Context.
\bibliography{anthology,eacl2021bib}
\bibliographystyle{acl_natbib}

\end{document}